\documentclass{article}

\usepackage[final]{neurips_2023}

\usepackage[utf8]{inputenc} 
\usepackage[T1]{fontenc}    
\usepackage{hyperref}       
\usepackage{url}            
\usepackage{booktabs}       
\usepackage{amsfonts}       
\usepackage{nicefrac}       
\usepackage{microtype}      
\usepackage{xcolor}         

\usepackage{graphicx}
\usepackage{amsmath}
\usepackage{algorithm}
\usepackage{algorithmic}
\usepackage{soul}
\DeclareMathOperator*{\argmax}{arg\,max}
\usepackage{longtable}
\usepackage{xcolor}

\newcommand{\ourmodel}{\textsc{PathFinder}}

\title{\ourmodel: Guided Search over Multi-Step Reasoning Paths}

\author{Olga Golovneva\thanks{corresponding author at \texttt{olggol@meta.com}}, Sean O'Brien, Ramakanth Pasunuru, Tianlu Wang, \\
\textbf{Luke Zettlemoyer, Maryam Fazel-Zarandi, Asli Celikyilmaz} \\
FAIR at Meta}

\begin{document}

\maketitle

\begin{abstract}
With recent advancements in large language models, methods like chain-of-thought prompting to elicit reasoning chains have been shown to improve results on reasoning tasks. However, tasks that require multiple steps of reasoning still pose significant challenges to state-of-the-art models. Drawing inspiration from the beam search algorithm, we propose \ourmodel, a tree-search-based reasoning path generation approach.
It enhances diverse branching and multi-hop reasoning through the integration of dynamic decoding, enabled by varying sampling methods and parameters.
Using constrained reasoning, \ourmodel{} integrates novel quality constraints, pruning, and exploration methods 
to enhance the efficiency and the quality of generation. Moreover, it includes scoring and ranking features to improve 
candidate selection.
Our approach outperforms competitive baselines on three complex arithmetic and commonsense reasoning tasks by $6\%$ on average.
Our model generalizes well to longer, unseen reasoning chains, reflecting similar complexities to beam search with large branching factors.
\end{abstract}

\section{Introduction}

Recent progress in large language models (LLMs) has led to a new era in machine reasoning, particularly through the use of prompting methods. These methods, such as chain-of-thought (CoT) \cite{wei2022cot}, scratchpads \cite{nye2021scratchpads}, least-to-most \cite{zhou2023leasttomost}, and program-aided language models (PAL) \cite{gao2023pal}, typically break down complex tasks into reasoning chains and have shown to improve model performance on tasks such as logical \cite{clark2020}, arithmetic \cite{cobbe2021training} and commonsense \cite{talmor2022} reasoning.

As tasks that require multiple steps of reasoning become more complex, LLMs begin to struggle with accumulating errors across multiple reasoning steps. 
Even more challenging is ensuring that each step in a chain is correctly evaluated and contributes positively to the overall reasoning chain and accuracy of the solution. To address these issues, recent work has implemented methods like self-consistency for majority voting \cite{wang2023selfconsistency}, diversifying prompts \cite{li2023making} and Python programs for more accurate reasoning generations \cite{gao2023pal}. Despite these improvements, the process of creating reasoning chains as a standard autoregressive process still faces challenges due to large search space, sub-optimal assessment and guidance of the reasoning process, especially in complex, multi-step tasks. 

\begin{figure*}[t]
\centering
\scalebox{0.8}{
    \includegraphics[trim={1.6cm 1.0cm 0.6cm 1.8cm},clip,width=1.\linewidth]{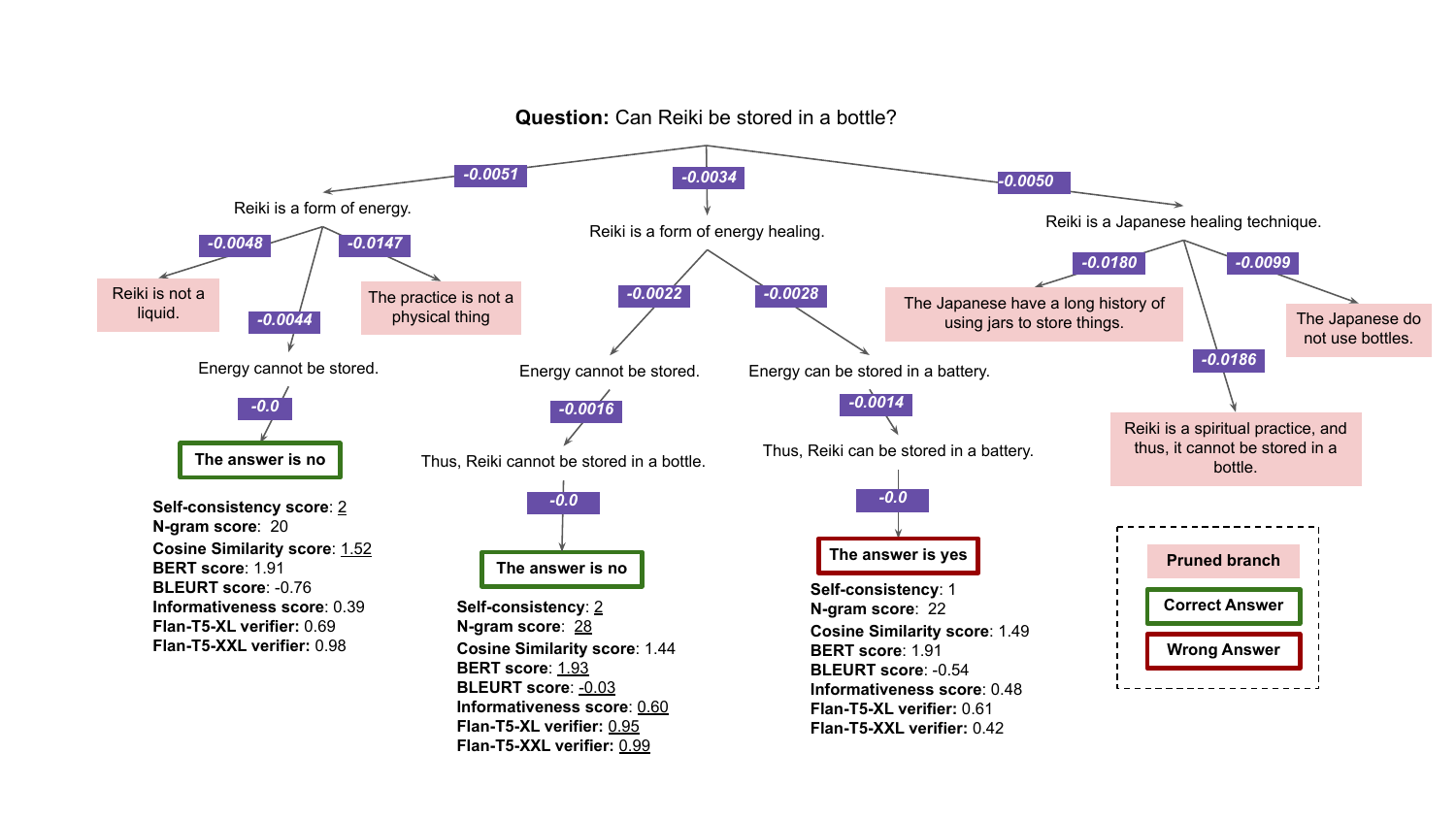}   }
    \caption{\footnotesize \ourmodel{} leverages step-level generic constrained decoding to guide step-by-step reasoning generations. In this example from the StrategyQA dataset, although the reasoning steps are close in n-grams, \ourmodel{} prunes less likely branches and chooses more informative ones that explain the question.
    Branching factor=3, buffer size=3. Numbers in purple rectangles are scores produced by the pruning function governed by Equation 2 using greedy step decoding ($\tau \xrightarrow{} 0$). Highest scores produced for each non-pruned branch are \underline{underlined}.
    Details on different candidate selection scores at the leaf of each reasoning path are provided in Section~\ref{ablation}.
    }
    \label{fig:tree_examples2}
\end{figure*}

In this work, we introduce \ourmodel, a decoding method designed for the generation and refinement of reasoning chains generated by LLMs. \ourmodel{} embodies our approach of dividing the reasoning decoding into two distinct tasks: \textit{candidate generation} and \textit{candidate selection}. For the candidate generation process, \ourmodel{} employs a tree-search-based method. It integrates a set of constraints to improve the quality of generated reasoning candidates, along with a pruning function for efficient computation and removes subpar candidates as shown in Figure~\ref{fig:tree_examples2}. Our model also incorporates an exploration factor to ensure the diversity of reasoning generations. For the candidate selection process, \ourmodel{} utilizes a set of novel similarity-based functions that we benchmark against existing LLM-based verifiers. This selection process allows for the selection of more accurate reasoning chains from the candidate pool, thereby refining the quality of the overall reasoning. We conduct extensive experiments across four generation tasks that necessitate multi-step reasoning. Using the small-size \textsc{LLaMA-7B} \citep{touvron2023llama} 
as a backbone language model, \ourmodel{} demonstrates substantial performance improvements across all tasks, highlighting its effectiveness in improving reasoning capabilities of language models.
We discuss related research in more detail in the Appendix~\ref{related}.

In summary, we develop \ourmodel, a new decoding method for effective generation of reasoning traces. Our algorithm is versatile, as it can be applied to a variety of multi-step reasoning generation tasks via decoding time constraints, reducing the need for tuning on costly labeled data. Our extensive experiments demonstrate its effectiveness on several complex tasks. 


\section{\ourmodel: Reasoning Decoder}
We describe \ourmodel, a tree-search-based reasoning path generation approach. We first introduce the decoding problem and then describe our approach which incorporates a two step decoding process: \textit{candidate} \textit{generation} and \textit{selection}.

\vspace{5pt}
\noindent \textbf{Decoding.} Sequence generation is a task of generating output sequence $\mathbf{y}$ (e.g., reasoning path) given input sequence $\mathbf{x}$ (e.g., question, prompt, etc.). In multi-step reasoning from LLMs, a sequential reasoning chain composed of \textit{T} steps is generated across multiple timesteps. We denote a reasoning chain as $\mathbf{y}=y_{1:T}$=$[y_1, y_2, \cdots, y_T]$, where each $y_t$ denotes a reasoning step of sequence of distinct tokens. A reasoning chain is autoregressively generated and the decoding consists of solving:
\begin{equation}
    \mathbf{y}_*=\argmax_{\mathbf{y}\in \mathcal{Y}} \log P(\mathbf{y}|\mathbf{x})
\end{equation}
where $\mathcal{Y}=\{\mathbf{y}_1, ..., \mathbf{y}_K\}$ is the set of all generated reasoning paths in response to the input $\mathbf{x}$ using generation model $P(\mathbf{y}|\mathbf{x})$. 

\vspace{5pt}
\noindent \textbf{Candidate Generation.} At the core of \ourmodel{} is a tree-search algorithm for long form reasoning generation, which generates multiple plausible generation paths. Unlike token-based beam search methods \cite{holtzman2019}, the branching in our approach occurs at the level of reasoning steps instead of individual tokens. This means that each reasoning step is regarded as a discrete node (see Figure~\ref{fig:tree_examples2}). The branching is influenced by varying among a fixed number of sampling parameters (e.g., top-\textit{k}, top-\textit{p}, temperature), allowing \ourmodel{} to explore various decoding methods and enabling search over different decoding strategies at each time step. This dynamic decoding property facilitates multi-hop reasoning, resulting in diverse branches.

To trade off quality against compute, we draw a number of candidates from each non-pruned leaf within the reasoning tree as our branching factor at every stage of the process. We continue to sample until all leaves have either reached termination points or have achieved a predefined maximum depth. 
To avoid over-generation of reasoning step branches, we also introduce a buffer size $b$, limiting the number of hypothesises stored for each context, and implement pruning methods. 
In particular, for each hypothesis reasoning step $y_t$= $[y_t^1,\cdots,y_t^N]$, where $y_t^i$ is the $i^{th}$ 
token in the sequence on length $N$, generated in response to the prompt $\mathbf{x}$, we prune branches based on the sequence scores
, normalized by the number of tokens:

\begin{equation}
    \pi(y_t) = \sum_{i} \log p_{\theta}(y_t^{i} | \mathbf{x}, y_t^{i<})/N^\lambda
\end{equation}
where $\lambda \in \mathbb{R}$ is a model-specific length penalty parameter
, and $p_{\theta}$ is a token generation model. Additionally, similar to ~\citet{xie2023decomposition}, we introduce step sampling temperature $\tau$ with annealing factor $\alpha$ used to decay temperature step-by-step as $\tau \xrightarrow{}\alpha\tau$ 
to add controlled variation in the branches, and sample according to the distribution:

\begin{equation}
    p(y_t) \propto \exp(\pi(y_t)/\tau)
\end{equation}

\vspace{5pt}
\noindent \textbf{Candidate Generation Constraints.} 
We enforce additional constraints on reasoning steps to reduce hallucinations. In particular, we force the model to re-generate a reasoning step if one of the following conditions is satisfied: (1) \textit{Repetition constraint}: generated step is similar to one of the previously generated steps or repeats context as determined by the cosine similarity metric. Cosine similarity is computed using the sentence embedding model \textit{all-mpnet-base-v2}~\cite{all-mpnet-base-v2}, and we force the model to re-generate if similarity value is greater than $0.9$; (2) \textit{Contradiction constraint}: generated step contradicts the context (i.e., question and previous steps) as determined by an entailment model. Specifically, we use the model proposed by~\citet{laurer2022less} to classify the step into one of three classes: \textit{entailment}, \textit{neutral}, and \textit{contradiction}, and force the model to re-generate if it belongs to the \textit{contradiction} class as determined by the entailment model.
If no new steps are generated after $n=2$ attempts, the branch is pruned. 

\vspace{5pt}
\noindent\textbf{Candidate Selection.} To select a final hypothesis out of a pool of candidates, we experiment with a number of scoring functions of the form:
\begin{equation}
    \mathbf{y}_* = \argmax_{\mathbf{y}_j \in \mathcal{Y}}{\sum_{\mathbf{y}_k \in \mathcal{Y}, \mathbf{y}_k \ne \mathbf{y}_j}{S(\mathbf{y}_j, \mathbf{y}_k)}}
\label{eq-selection}
\end{equation}
where the number of candidate reasoning chains in the pool $\mathcal{Y}$ is limited by the buffer size ($K\le b$), and $S$ is a similarity function.
The intuition is similar to \textit{self-consistency} \citep{wang2022self} or \textit{wisdom of the crowd} \citep{suzgun2022follow}, in the assumption that a solution following from more diverse, generated reasoning chains majority is more likely to be the correct one. In fact, our results support the use of an \textit{N-gram}-based similarity metric
. Specifically, if $g_j$ is a set of n-grams for the hypothesis $\mathbf{y}_j$, the \textit{N-gram} similarity function is defined as the number of common n-grams as follows:
\begin{equation}
    S(\mathbf{y}_j, \mathbf{y}_k) = |g_j \cap g_k| \label{eq-ngrams}
\end{equation}

Candidate selection is a critical component of \ourmodel. Common techniques are using scorer functions and verifier models. Scorer functions~\cite{suzgun2022follow, prasad2023receval} 
help rank fixed set of candidate generations and guide the selection of the final prediction based on some property of the generated text, such as similarity. On the other hand verifier models \cite{li2023making} 
use external models to explicitly evaluate the correctness of the produced hypothesis, and rank generations based on the faithfulness score. 
We validate \ourmodel{} against verifier models and different similarity-based scorers in our ablation studies in Section~\ref{ablation}. We note that the usage of a suitable scoring function is preferred over a verifier model as it would improve runtime and reduce memory load. 

\begin{table*}[t]
\centering
\small
\begin{tabular}{lccc}
\hline
\textbf{Model} & \textbf{GSM8K} & \textbf{StrategyQA} & \textbf{CSQA} \\
\hline
\textsc{GPT-6.7B} & 2.4 & 50.0 & 24.0\\
\textsc{Minerva-8B} & \textbf{16.2} & - & -\\
\textsc{Llama-7B} & 11.0 & 61.1* & 43.3* \\
\textsc{Llama-7B} (self consistency) & 15.3* & \underline{64.8*} & 46.9*\\
\textsc{Flan-T5-XL (3B)} & 13.5 & \textbf{73.4*} & \textbf{85.4*}\\
\hline
\ourmodel~ (\textsc{LLaMa-7B}, N-gram) & 11.3 & 59.0 & 50.0\\
\ourmodel~ (\textsc{LLaMa-7B}, \textsc{Flan-T5-XL}) & 11.7 & 60.8 & 55.1\\
\ourmodel~ (\textsc{LLaMa-7B}, \textsc{text-davinci-003}) & \underline{15.4} & 61.7 & \underline{56.3}\\
\hline
\end{tabular}
\caption{Performance of different models on four reasoning benchmark datasets measured with accuracy. Best numbers are \textbf{bolded} among models and the second best numbers are \underline{underlined}. Numbers with an asterisk* are from our evaluations using greedy decoding and CoT prompts provided in Appendix~\ref{app:exp_setup}. For self-consistency scores, we marginalize answer across 16 reasoning chains sampled with temperature $T=1.0$, \textit{top-k} ($k=40$) and \textit{top-p} ($p=0.5$). We note that \textsc{FLAN-T5} is finetuned on data from both the \textsc{CSQA} and \textsc{GSM8K} datasets and thus will have somewhat inflated performance in comparison to comparably-sized models not trained on these tasks.
}
\label{tab:results}
\end{table*}

\section{Experiments: Reasoning Generation}
\label{exp}
\noindent\textbf{Datasets.} We conduct experiments on various benchmark datasets that require complex reasoning skills to reach the final answer: (1) \textsc{GSM8K} \citep{cobbe2021training}, an arithmetic reasoning dataset of 8.5K linguistically diverse grade school math word problems; (2) \textsc{StrategyQA} \citep{geva2021did}, a commonsense reasoning dataset of 2,780 questions, annotated with their decomposition and
per-step evidence; (3) \textsc{CSQA} \citep{talmor2018commonsenseqa}, a multiple-choice commonsense reasoning dataset of 12,102 questions with one correct answer and four distractor answers.

\vspace{5pt}
\noindent\textbf{Backbone LLMs for \ourmodel.} We select two widely-used open-sourced models to generate and evaluate chains of reasoning: \textsc{LLaMa-7B} \citep{touvron2023llama} and \textsc{Flan-T5-XL (3B)}
\citep{chung2022scaling}. We prompt 
\textsc{LLaMa-7B} model
with chain-of-thought examples \citep{wei2022cot} to generate reasoning steps along with the final answers. We provide specific parameter values and prompt sequences in Appendix~\ref{app:exp_setup}. We also experiment with different methods for candidate selection. 
In particular, we report results using the following setups: (1) \ourmodel{} (\textsc{LLaMa-7B}, N-gram): uses \textsc{LLaMa-7B} model for text generation, and tri-gram similarity for candidate selection; (2) \ourmodel{} (\textsc{LLaMa-7B}, \textsc{FLAN-T5-XL}): uses \textsc{LLaMa-7B} model for text generation, and \textsc{FLAN-T5-XL} verifier model for candidate selection; (3) \ourmodel{} (\textsc{LLaMa-7B}, \textsc{text-davinci-003}) uses LLaMa-7B model for text generation, and  \textsc{text-davinci-003} verifier model from the family of \textsc{GPT-3.5} models for candidate selection.

\noindent\textbf{Baselines.} We benchmark our approach against leading best models with reported results in the literature, ensuring the model sizes are comparable for fair evaluation\footnote{We also attempted to generate results using the self-evaluation guided decoding method described in \cite{xie2023decomposition} for the LLaMa-7B model. However, the process was slow and computationally expensive: Reducing the beam size to 2 and number of samples at each step to 4, this method took over 4 days to run and achieved an accuracy of 9.5 on GSM8K and 58.3 on StrategyQA.}. Specifically we compare against \textsc{GPT-6.7B} \cite{wei2022cot}, \textsc{LLaMa-7B} \citep{touvron2023llama}, \textsc{Flan-T5-XL (3B)}
\citep{fu2023chain}, 
and Minerva-8B \cite{lewkowycz2022solving}. Reported results represent evaluation results on generations produced with CoT prompting and greedy token-level decoding. We also include our own evaluations on a few tasks that to the best of our knowledge are missing in the literature using greedy decoding and prompts provided in Appendix~\ref{app:exp_setup}.

\vspace{5pt}
\noindent \textbf{Results.} Table~\ref{tab:results} compares different LLMs with different decoding methods showing answer accuracy as the evaluation metric. \ourmodel~improves baseline performance on all selected reasoning tasks by $6\%$ on average, but lacks behind the base model with self-consistency applied on \textsc{StrategyQA} dataset by $3\%$. Even simple N-gram-based similarity metric allows to select better paths leading to model improvements with respect to baseline on \textsc{GSM8K} and \textsc{CSQA} datasets.
 We note that \textsc{Flan-T5-XL} verifier significantly improves performance on \textsc{CSQA} task, but not that much on the others. This is likely due to the fact that is was trained on this task, while other tasks are significantly harder to evaluate (\textsc{GSM8K}), or not familiar to the model (\textsc{StrategyQA}). While overall \textsc{text-davinci-003} verifier shows better performance, there is a trade-off between the amount of resources needed to run \textsc{GPT3.5}-based evaluations and improvements in performance it could give;
Figure~\ref{fig:tree_examples2} shows an example of the generated tree for one of the \textsc{StrategyQA} questions. It showcases how \ourmodel{} that although the reasoning steps are close in N-grams, \ourmodel{} prunes less likely branches and chooses more informative ones that explain the question. Finally, the N-gram scorer selects the correct answer by selecting the branch with higher n-gram similarity to other branches.

\section{Ablation Study}
\label{ablation}
\begin{figure}
    \includegraphics[width=.5\linewidth]{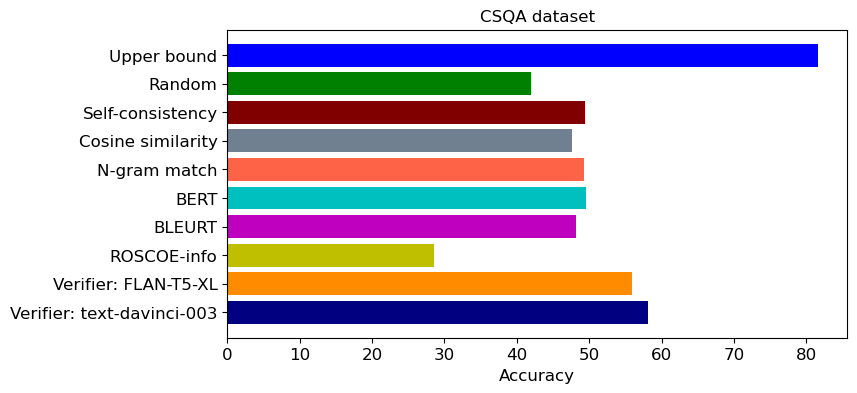} 
    \includegraphics[width=.5\linewidth]{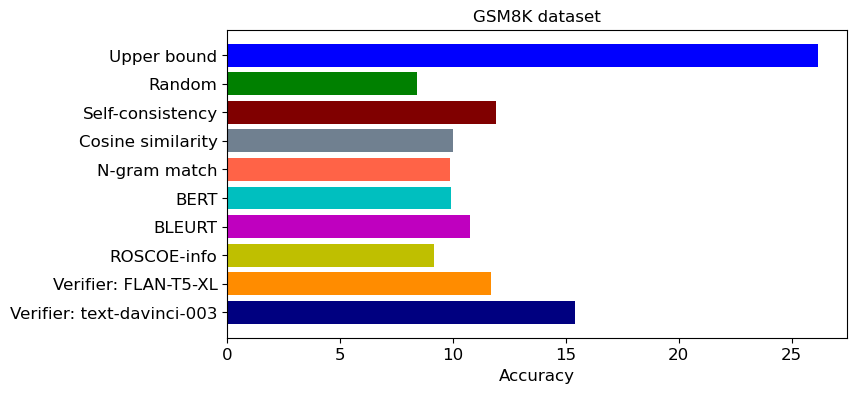}   
    \caption{\footnotesize The impact of different similarity-based scoring functions and verifier models (vertical axis) on the accuracy score (horizontal axis) of \ourmodel{}, utilizing \textsc{Llama-7B} as backbone LLMs, on the \textsc{CSQA} and \textsc{GSM8K} datasets. We use buffer size 128 with branching factor of 8 for \textsc{CSQA}, and buffer size 16 with branching factor of 4 for \textsc{GSM8K} dataset. Scoring functions score and rank hypothesises based on similarity metrics, while verifier model ranks hypothesises based on the generated faithfulness score.
    }
    \label{fig:cost_funct}
\end{figure}

\textbf{How do various candidate selection strategies impact the overall performance?} 
In this section, we evaluate our model using various scorer functions and verifier models, which rank a fixed set of candidates and select the highest-scoring one as the final prediction. We present an upper bound accuracy, which is an accuracy of the "perfect" scorer that would select the correct final answer if present in candidate pool, and contrast the \textit{N-gram}-based scorer with several alternative approaches: \textit{Self-Consistency scorer}, \textit{Cosine Similarity scorer}, \textit{\textsc{BERT}  and \textsc{BLEURT} scorers}, \textit{Informativeness scorer}, and \textit{Verifier models}. We provide more details on scorer functions construction and ranking methods in Appendix~\ref{app:exp_setup} and Appendix~\ref{app:verifier}. We summarize the results in Figure~\ref{fig:cost_funct}. All scorers outperform random selection, with \textsc{text-davinci-003} results in highest accuracy score of 58.1. At the same time we want to emphasize the gap between the upper bound accuracy 
and the final accuracy when a scorer function or verifier model is used to rank and select best hypothesis, which clearly shows that the right choice of the candidate selection strategy can significantly boost performance further.

\begin{figure}[t]
    \centering
    \includegraphics[width=.45\linewidth]{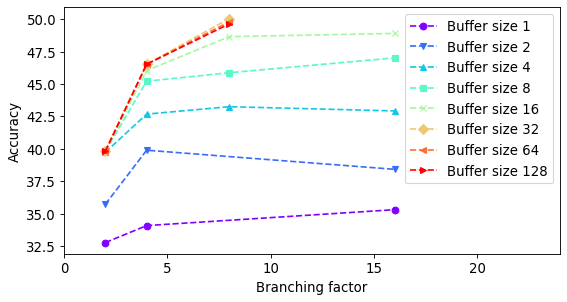} 
    \includegraphics[width=.45\linewidth]{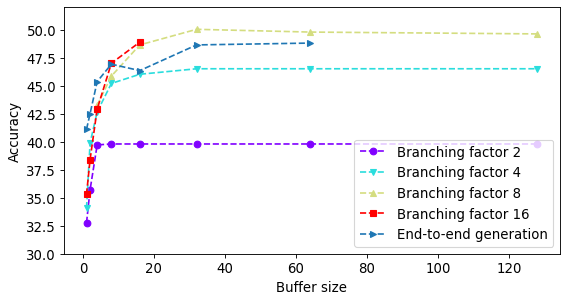} 
    \caption{\footnotesize  \ourmodel{} (\textsc{LLaMa-7B}, N-gram) accuracy scores on \textsc{CSQA} dataset as a function of barnching factor (left) and buffer size (right). On the right figure we also include the results where we use \textsc{LLaMa-7B} to generate the whole reasoning path for each candidate, and then apply tri-gram scorer model on the number of generated candidates corresponding to the buffer size value (denoted as \textit{end-to-end generation}). Decoding process is sensitive to the amount of noise, which results in the existence of an optimal branching factor that maximizes the final accuracy. For each branching factor there is a limit on the maximal number of diverse branches the model can generate, and increasing buffer size after this point does not result in bigger trees, and we observe a platoe in terms of accuracy scores.
    }
    \label{fig:branching_factor}
\end{figure}

\noindent\textbf{How does the branching factor affect performance?} The tree branching factor together with the pruning function significantly influences the diversity of candidates. Intuitively, generating more candidates increases the likelihood of producing at least one correct generation. However, as our scoring models are not perfect, a high volume of noisy candidates could confuse them and escalate the rate of false positives. We 
asses the \textit{N-gram} scorer performance to comprehend the scorer sensitivity to noise. Figure~\ref{fig:branching_factor} indicates an optimal branching factor for each buffer size, which supports our hypothesis regarding the scorer function's sensitivity to noise levels. Thus, for tree-search-based step-level decoding it is important to find an optimal value of the branching factor to balance between the diversity of the candidates and the amount of noise. Similar phenomenon was previously observed for beam search token-level decoding, where increasing the number of decoding candidates past a certain point leads to the worse generation quality~\citep{yang-etal-2018-breaking, koehn-knowles-2017-six}. 

\noindent\textbf{Does an increased buffer size consistently improve performance?} To answer this question we empirically investigate the 
\textit{N-gram} scorer performance.
The results summarized in Figure~\ref{fig:branching_factor} reveal that for small branching factors considered in this study, the accuracy score platoes. Beyond a certain point, increasing the buffer size does not yield more generations because they are limited by the branching factor and generation constrains. In fact, for \textsc{CSQA} experiments shown in Figure~\ref{fig:branching_factor}, at branching factor 8 the average number of hypothesis candidates almost does not change after buffer size 32, and is around 22. The platoe point shifts higher with the increase in the number of generations per node. Throughout our experiments, we observed a consistent improvement in optimal performance with the an increased buffer size. However, our observations are limited by available computational resources, suggesting that behavior might vary for extreme branching factors. 

\noindent\textbf{Do we actually need trees?} In Figure~\ref{fig:branching_factor}, along with the tree-search-based results, we also report end-to-end generation performance, i.e., at candidate generation stage, instead of creating a tree we prompt the model to generate a set of reasoning chains, and then apply our candidate selection process. We observe that a sufficient diversity in reasoning steps is necessary for the tree-search-based approach to outperform the end-to-end generation method. Specifically, for the \textsc{LLaMa-7B} model, the tree-search generation approach only outperforms end-to-end generation at a branching factor of 8 or above. Therefore, while tree-search generation has its advantages, it's effectiveness in comparison to end-to-end generation is largely dependent on sufficient diversity in reasoning steps and a relatively high branching factor.   



\begin{table}[t]
\centering
\small
\begin{tabular}{lccccc}
\hline
& & \multicolumn{4}{c}{\textbf{Branching Factor}} \\
\textbf{Dataset} & \textbf{End-to-end} & \textbf{2} & \textbf{4} & \textbf{8} & \textbf{16}\\
         \hline
         \textbf{GSM8K} 
& 0.8 & 0.5 & 1.6 & 5.3 & - \\
        \textbf{StrategyQA} 
& 0.2 & 0.2 & 0.8 & 1.6 & 3.3 \\
        \textbf{CSQA} 
& 0.07 & 0.06 & 0.2 & 0.6 & 1.4 \\
\hline
\end{tabular}
\caption{\footnotesize Average GPU-hours required \textsc{LlaMa-7B} model to generate a reasoning tree for different branching factors $br$. Batch of 16 samples was used in all datasets. All numbers are calculated for buffer size 8, we also report average GPU-hours for end-to-end generation and sampling size 8.
    }   
\label{tab:runtime}
\end{table}

\noindent\textbf{Computational complexity} The benefits gained from this approach come at a high computational cost. Assuming a full buffer of $c$ candidates after a few reasoning steps, we generate $cb$ candidate new steps before selection and pruning at each step. If the original reasoning chain required $T$ tokens to generate, \ourmodel{} requires $O(bcT)$ tokens. This can be compared against self-consistency with $k$ paths, which requires $O(kT)$ tokens, but can be more easily parallelized with memory split across devices. In practice, we see that, while a simple reasoning path takes under 1 GPU-hour to generate (Table~\ref{tab:runtime}), it takes several GPU-hours to generate reasoning tree that would outperform end-to-end generation for complex reasoning tasks. With more effective pruning functions and scorers, we should be able to realize gains with fewer sampled paths. Because our framework is agnostic to the choice of either, we anticipate that better models will allow for a lower branching factor and buffer size, making the method more computationally feasible.

\section{Conclusion}
We proposed \ourmodel{}, a novel decoding method that optimizes reasoning chain generation in large langauge models. We demonstrate notable performance improvements on four multi-step reasoning tasks, emphasizing its versatility and effectiveness. Our approach overcomes traditional limitations, enhancing reasoning capabilities and opening doors for future research.



\bibliography{anthology,custom}
\bibliographystyle{acl_natbib}

\appendix

\section{Limitations}
Our study is limited by the number of models and tasks we used to empirically support the proposed approach. Although \ourmodel{} outperforms other baselines on selected tasks, it comes with significant increase in computational complexity. To be able to efficiently use tree-search-based step-level decoding, we would need to develop more effective sampling and scoring techniques that will allow us to achieve high-quality results faster and at lower computational costs.

\section{Ethics Statement}
Our method, \ourmodel{}, improves text generation, specifically focused on step-by-step rationale generation from large language models. Thus, it inherits the potential benefits and risks associated with text generation applications \cite{NEURIPS2020_1457c0d6}. By imposing logical constraints on text generation, we aim to enhance control, consistency, and accuracy, specifically in tasks that require step-by-step reasoning, such as arithmetic reasoning. We would like to note that any language model, even under constraints could potentially be exploited to produce biased, or offensive narratives \cite{mcguffie2020}. For in-depth exploration of these risks, we direct the reader to the analysis presented in \citep{bender2021}.

\section{Related work}
\label{related}

\vspace{5pt}
\noindent\textbf{Decoding strategies for text generation.} 
These methods present a continual trade-off between quality and diversity. Traditional deterministic methods such as greedy decoding and beam search \cite{jurafsky2009,graves2012} offer high-quality results but can lack diversity and are prone to degeneration. Truncation-based sampling methods such as temperature sampling, top-\textit{k} sampling, top-\textit{p} sampling and locally typical sampling have been used to balance this trade-off~\cite{holtzman2019,meister2023}. The advent of autoregressive LLMs like GPT has spurred numerous works focusing on various factors such as diversity~\cite{ippolito2019comparison}, fluency~\cite{holtzman2019}, and constraint satisfaction~\cite{anderson-etal-2017-guided,miao2019cgmh,welleck2019non,lu-etal-2021-neurologic} in decoding strategies. Constrained decoding methods have seen enhancements like grid beam search~\cite{anderson-etal-2017-guided} and constrained beam search~\cite{hokamp-liu-2017-lexically} that aim at satisfying lexical constraints during generation. Other works such as Metropolis-Hastings sampling-based conditional generation~\cite{miao2019cgmh} and tree-based constrained text generation~\cite{welleck2019non} seek to address the mismatch between monotonic decoding and satisfying constraints. Contrary to these strategies which often struggle with the balance between quality and diversity, \ourmodel{} focuses primarily on reasoning tasks and not open-ended text generation, operating on reasoning steps rather than on individual tokens. By separating out the steps of tree-search-based generation and similarity-based selection, our approach generates diverse reasoning chains and refines them for optimal quality.

\vspace{5pt}
\noindent\textbf{Advanced CoT Strategies and Self Consistency.} It has been shown that aggregating from diverse CoTs (i.e., multiple reasoning paths for each problem) can effectively enhance end-task performance~\cite{wei2022cot}. Recent works such as self-consistency~\cite{wang2023selfconsistency} and crowd sampling~\cite{suzgun2022follow} generate multiple reasoning paths and try to find a consensus among the derived answers. Self-consistency has significantly boosted performance in CoT reasoning, even in tasks where CoT prompting traditionally harms performance; crowd sampling has shown gains in non-reasoning tasks like summarization. Our approach bears resemblance to these, but differs in its use of a tree search for candidate generation, its operation on the step level rather than on the full generations, and its use of a novel similarity function. 

Other approaches offload some portion of problem-solving to external tools, like training a model to use tools as in Toolformer~\cite{schick2023toolformer} or prompting a model to solve a problem with a code interpreter as in PAL~\cite{gao2023pal}.
Our approach does not require access to external tools, although it does not prohibit their use. Further, unlike Toolformer our method is training-free.

Our work parallels the study of by~\citeauthor{xie2023decomposition}, which presents a self-evaluation guided stochastic beam search for multi-step reasoning. Their method employs beam search decoding tailored to intermediate steps and guides the searching process by controlling the error of each reasoning step to prevent potential error accumulation. Another approach for step selection and evaluation was recently developed by \cite{hao2023reasoning} and \cite{shridhar2023art}, that relies on the world model to ask, refine, and evaluate generated steps. 

\section{Experimental setup}
\label{app:exp_setup}

\textbf{Inference parameters.} To run experiments with \ourmodel{} we used \textsc{LLaMa-7B}. For step-by-step generations, we applied temperature token sampling with $T=1.0$, with additional \textit{top-k} ($k=40$) and \textit{top-p} ($p=0.5$) truncation to increase the diversity of the samples during tree generation. For end-to-end generation we applied greedy decoding. We fixed the maximum generation length at 128 tokens per step for tree generation, and 512 for the full reasoning generation. We run experiments on 8 GPUs with batch size 16.

\textbf{Prompt construction.} All prompts used for hypothesis generations are listed in Table~\ref{tab:prompts}. In particular, for \textsc{GSM8K} dataset we follow prompts from~\citet{touvron2023llama}, for \textsc{StrategyQA} and \textsc{CSQA} datasets we follow~\citet{wei2022cot}. 


\textbf{Pruning.} In main experiments we applied annealing factor $\alpha=0.5$, and used step sampling temperature $\tau=1.0$.  $\tau=0$ corresponds to the maximum likelihood sampling, while $\tau \xrightarrow{}\inf$ corresponds to the uniform sampling. This setup allows for higher variation of steps in the beginning of generation, and becomes more strict with depth. We have experimented removing annealing factor and varying $\tau=\{0, 0.5, 1.0, 16\}$ on the train partition of \textsc{GSM8K} dataset, and found the optimal performance at $\tau=1$ with annealing.

\textbf{Scoring functions.} We contrast the \textit{N-gram}-based scorer with several alternative approaches: 
\begin{itemize}
    \item \textit{Self-Consistency scorer}: The final answer is determined by marginalizing out the sampled reasoning paths to find the most consistent answer in the final answer set~\citep{wang2022self}. This method does not take into account reasoning chains, so we modify Equation~\ref{eq-selection} as $\textbf{y}_* = \argmax_{y_j \in \mathcal{Y}}{n_{a_j}}$, where $\mathcal{A}=\{a_1, ..., a_b\}$ are all generated answers extracted from corresponding hypothesises $\mathcal{Y}$, and each answer $a_j$ appears $n_{a_j}$ times in $\mathcal{A}$.
\item  \textit{Cosine Similarity scorer}: The final answer is selected by marginalizing out the total cosine similarity of the reasoning chains, so we modify the Equation~\ref{eq-ngrams} as $S(\mathbf{y}_j, \mathbf{y}_k) = \cos(\mathbf{e}_j, \mathbf{e}_k)$, where $\mathbf{e}_j$ is the embedding of the hypothesis reasoning path $\mathbf{y}_j$ as determined by the \textit{all-mpnet-base-v2}~\cite{all-mpnet-base-v2} sentence embedding model. 
\item  \textit{\textsc{BERT}  and \textsc{BLEURT} scorers}: Following \textit{wisdom of the crowd} work~\cite{suzgun2022follow}, we try \textsc{BLEURT}~\citep{sellam-etal-2020-bleurt} and \textsc{BERTScore}~\citep{zhang2019bertscore} metrics as similarity function $S$ in Equation~\ref{eq-selection}. 
\item  \textit{Informativeness scorer}: We select the final hypothesis based on the amount of information shared between the source context $\mathbf{c}$ and the reasoning paths, measured through mutual alignment. Specifically, we use the \textit{Info-chain} score defined as $I(\mathbf{y}_j, \mathbf{c})$ in~\citep{golovneva2022roscoe}, and revise Equation~\ref{eq-selection} for this scorer function as $\mathbf{y}_* = \argmax_{\mathbf{y}_j \in \mathcal{Y}}I(\mathbf{y}_j, \mathbf{c})$.
\item  \textit{Verifier models}: We select the final hypothesis reasoning path and the answer based on the score provided by a pre-trained verifier model. We use \textsc{Flan-T5-XL}  and \textsc{text-davinci-003} models to rank the hypothesises and select the one ranked at the top. To score the hypothesises, models are prompted to evaluate the correctness of the reasoning path, then hypothesises are ranked based on returned faithfulness score. We provide more details of the ranking method in Appendix~\ref{app:verifier}.
\end{itemize}

\begin{small}
\begin{longtable}{|p{14cm}|}
\hline
\textbf{Few-shot prompts used for GSM8K dataset} \\
\hline
Answer these questions:\\
\\
\textbf{Q}: There are 15 trees in the grove. Grove workers will plant trees in the grove today. After they are done, there will be 21 trees. How many trees did the grove workers plant today?\\
\textbf{A}: There are 15 trees originally. Then there were 21 trees after some more were planted. So there must have been 21 - 15 = 6. The answer is 6. \\
\\
\textbf{Q}: If there are 3 cars in the parking lot and 2 more cars arrive, how many cars are in the parking lot?\\\textbf{A}: There are originally 3 cars. 2 more cars arrive. 3 + 2 = 5. The answer is 5. \\ \\

\textbf{Q}: Leah had 32 chocolates and her sister had 42. If they ate 35, how many pieces do they have left in total? \\\textbf{A}: Originally, Leah had 32 chocolates. Her sister had 42. So in total they had 32 + 42 = 74. After eating 35, they had 74 - 35 = 39. The answer is 39. \\ \\

\textbf{Q}: Jason had 20 lollipops. He gave Denny some lollipops. Now Jason has 12 lollipops. How many lollipops did Jason give to Denny? \\\textbf{A}: Jason started with 20 lollipops. Then he had 12 after giving some to Denny. So he gave Denny 20 - 12 = 8. The answer is 8. \\ \\
\textbf{Q}: Shawn has five toys. For Christmas, he got two toys each from his mom and dad. How many toys does he have now? \\\textbf{A}: Shawn started with 5 toys. If he got 2 toys each from his mom and dad, then that is 4 more toys. 5 + 4 = 9. The answer is 9. \\ \\
\textbf{Q}: There were nine computers in the server room. Five more computers were installed each day, from monday to thursday. How many computers are now in the server room? \\\textbf{A}: There were originally 9 computers. For each of 4 days, 5 more computers were added. So 5 * 4 = 20 computers were added. 9 + 20 is 29. The answer is 29. \\ \\
\textbf{Q}: Michael had 58 golf balls. On tuesday, he lost 23 golf balls. On wednesday, he lost 2 more. How many golf balls did he have at the end of wednesday?\\\textbf{A}: Michael started with 58 golf balls. After losing 23 on tuesday, he had 58 - 23 = 35. After losing 2 more, he had 35 - 2 = 33 golf balls. The answer is 33. \\ \\\textbf{Q}: Olivia has \$23. She bought five bagels for \$3 each. How much money does she have left?\\\textbf{A}: Olivia had 23 dollars. 5 bagels for 3 dollars each will be 5 x 3 = 15 dollars. So she has 23 - 15 dollars left. 23 - 15 is 8. The answer is 8.\\
\\\hline 
\textbf{Few-shot prompts used for StrategyQA dataset} \\
\hline \\
Answer these questions:\\ \\\textbf{Q}: Do hamsters provide food for any animals? \\\textbf{A}: Hamsters are prey animals. Prey are food for predators. Thus, hamsters provide food for some animals. The answer is yes. 
\\ \\\textbf{Q}: Could Brooke Shields succeed at University of Pennsylvania?\\ \textbf{A}: Brooke Shields went to Princeton University. Princeton University is about as academically rigorous as the University of Pennsylvania. Thus, Brooke Shields could also succeed at the University of Pennsylvania. The answer is yes. \\ \\\textbf{Q}: Yes or no: Hydrogen’s atomic number squared exceeds number of Spice Girls?\\ \textbf{A}: Hydrogen has an atomic number of 1. 1 squared is 1. There are 5 Spice Girls. Thus, Hydrogen’s atomic number squared is less than 5. The answer is no. \\ \\\textbf{Q}: Yes or no: Is it common to see frost during some college commencements?\\ \textbf{A}: College commencement ceremonies can happen in December, May, and June. December is in the winter, so there can be frost. Thus, there could be frost at some commencements. The answer is yes. \\ \\\textbf{Q}: Yes or no: Could a llama birth twice during War in Vietnam (1945-46)?\\ \textbf{A}: The War in Vietnam was 6 months. The gestation period for a llama is 11 months, which is more than 6 months. Thus, a llama could not give birth twice during the War in Vietnam. The answer is no. \\ \\\textbf{Q}: Yes or no: Would a pear sink in water?\\ \textbf{A}: The density of a pear is about 0.6g/cm3, which is less than water. Objects less dense than water float. Thus, a pear would float. The answer is no.\\
\\\hline
\textbf{Few-shot prompts used for CSQA dataset} \\
\hline\\
Answer these questions:\\ \\\textbf{Q}: What do people use to absorb extra ink from a fountain pen?\\ \textbf{Answer Choices}:\\ (A) shirt pocket\\ (B) calligrapher’s hand\\ (C) inkwell\\ (D) desk drawer\\ (E) blotter\\ \textbf{A}: The answer must be an item that can absorb ink. Of the above choices, only blotters are used to absorb ink. The answer is E. \\ \\Q: What home entertainment equipment requires cable?\\ \textbf{Answer Choices}:\\ (A) radio shack\\ (B) substation\\ (C) television\\ (D) cabinet\\ \textbf{A}: The answer must require cable. Of the above choices, only television requires cable. The answer is C. \\ \\Q: The fox walked from the city into the forest, what was it looking for?\\ \textbf{Answer Choices}:\\ (A) pretty flowers\\ (B) hen house\\ (C) natural habitat\\ (D) storybook\\ \textbf{A}: The answer must be something in the forest. Of the above choices, only natural habitat is in the forest. The answer is B. \\ \\Q: Sammy wanted to go to where the people were. Where might he go?\\ \textbf{Answer Choices}:\\ (A) populated areas\\ (B) race track\\ (C) desert\\ (D) apartment\\ (E) roadblock\\ \textbf{A}: The answer must be a place with a lot of people. Of the above choices, only populated areas have a lot of people. The answer is A. \\ \\Q: Where do you put your grapes just before checking out?\\ \textbf{Answer Choices}:\\ (A) mouth\\ (B) grocery cart\\ (C) super market\\ (D) fruit basket\\ (E) fruit market\\ \textbf{A}: The answer should be the place where grocery items are placed before checking out. Of the above choices, grocery cart makes the most sense for holding grocery items. The answer is B. \\ \\Q: Google Maps and other highway and street GPS services have replaced what?\\ \textbf{\textbf{Answer Choices}:}:\\ (A) united states\\ (B) mexico\\ (C) countryside\\ (D) atlas\\ \textbf{A}: The answer must be something that used to do what Google Maps and GPS services do, which is to give directions. Of the above choices, only atlases are used to give directions. The answer is D. \\ \\Q: Before getting a divorce, what did the wife feel who was doing all the work?\\ \textbf{\textbf{Answer Choices}:}:\\ (A) harder\\ (B) anguish\\ (C) bitterness\\ (D) tears\\ (E) sadness\\ \textbf{A}: The answer should be the feeling of someone getting divorced who was doing all the work. Of the above choices, the closest feeling is bitterness. The answer is C.\\
\hline
\caption{\footnotesize Prompts used for hypothesis generation per dataset.}   
\label{tab:prompts}
\end{longtable}
\end{small}

\section{Verifier model}
\label{app:verifier}

To rank the generated hypothesises we can use pre-trained LLMs to that known to be well calibrated with respect to the True/False questions, and thus were used for self-evaluation~\citep{kadavath2022language}. We adopt this approach and extend it to using any external LLM model by prompting it with 5 shots of multiple-choice questions to identify if the generated reasoning is (\textsc{A}) correct or (\textsc{B}) incorrect. We use \textsc{FLAN-T5} and \textsc{text-davinci-003} models for evaluation, and extract the probability of the reasoning being correct as a faithfulness score. We note that \textsc{FLAN-T5} is finetuned on the training partitions of \textsc{CSQA} and \textsc{GSM8K}, and thus will have somewhat inflated performance in comparison to comparably sized models not trained on these tasks. We follow~\cite{xie2023decomposition} and use the probability of the option \textsc{A} as a score to rank and select generations. 

\end{document}